\documentclass{article}
\usepackage{arxiv}      

\usepackage[T1]{fontenc}
\usepackage{newtxtext}
\usepackage{newtxmath}

\newif\ifshowlogo
\showlogotrue

\usepackage{amsmath}

\usepackage{amsthm}

\usepackage{booktabs}
\usepackage{array}
\usepackage{ragged2e} 
\usepackage{tabularx}
\usepackage{longtable}

\newcolumntype{P}[1]{>{\RaggedRight\arraybackslash}p{#1}}
\usepackage{graphicx}
\usepackage{caption}
\usepackage{multirow}
\usepackage{xcolor}
\usepackage{xurl} 

\usepackage{eso-pic}
\newif\ifshowarxivstamp
\showarxivstampfalse 
\newcommand{\arxivstamptext}{arXiv:XXXX.XXXXX [cs.AI]\\\today}
\AddToShipoutPictureBG{%
  \ifshowarxivstamp
    \ifnum\value{page}=1\relax
      \AtPageLowerLeft{%
        \raisebox{3cm}[0pt][0pt]{%
          \hspace{0.25cm}%
          \rotatebox{90}{\footnotesize\color{black!60}\arxivstamptext}%
        }%
      }%
    \fi
  \fi
}

\usepackage[numbers,sort&compress]{natbib}
\usepackage[hidelinks,unicode]{hyperref}
\hypersetup{
  pdftitle  = {From Frame-Level Recognition to Event-Level Confirmation},
  pdfauthor = {M. Meng and Y. Zhang}
}

\newcommand{\rAF}{\texttt{rAF}}

\title{From Frame-Level Recognition to Event-Level Confirmation:\\
       Repair Traces and Runtime Failure Analysis of\\
       Public-Space Gesture Interaction}

\renewcommand{\shorttitle}{Repair Traces of Public-Space Gesture Interaction}

\author{%
  {\large\bfseries\fontseries{bx}\selectfont
    M. Meng\thanks{Corresponding author: \texttt{m@9zzg.com}},
    Yansong Zhang
  }\\[2pt]
  {\normalsize
    Shenzhen The Nine's Light Technology Co., Ltd., Shenzhen, Guangdong, P.R.~China
  }\\[2pt]
  {\normalsize
    \texttt{m@9zzg.com}\quad \url{https://9zzg.com}
  }
}

\date{May 4, 2026}

\begin{document}

%
\ifshowlogo
  \vspace*{-6pt}
  \noindent
  \IfFileExists{figures/moduoduo-logo.pdf}{%
    \includegraphics[height=14pt]{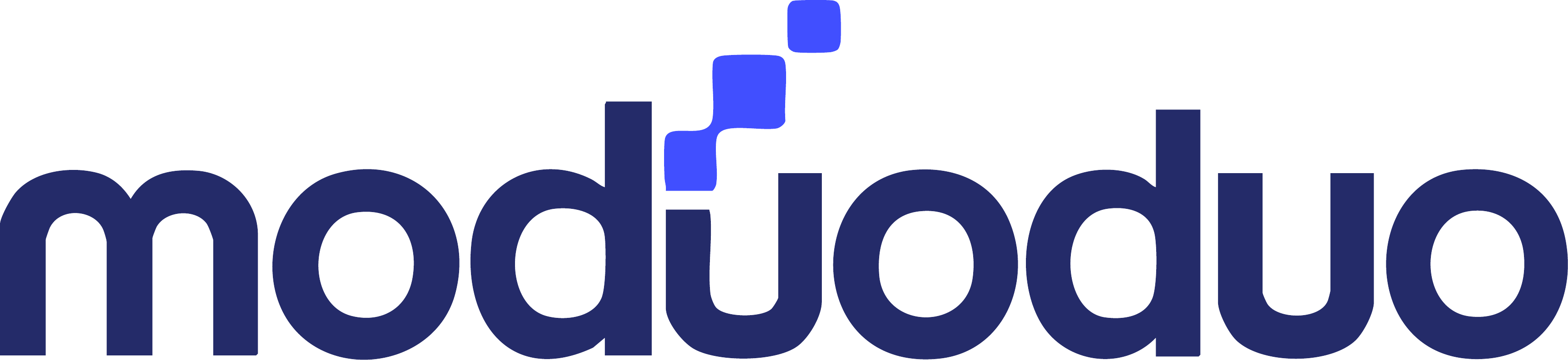}%
  }{%
    \includegraphics[height=14pt]{figures/moduoduo-logo.png}%
  }%
  \vspace{2pt}
\fi

\maketitle

\begin{abstract}
Deploying public-space gesture interaction is challenging not only because
single-frame recognition must be accurate, but also because low-level
recognition outputs must be organized into interaction events that users
can perceive and understand. In scenic areas, exhibition spaces, and public
service terminals, user-visible failures often appear as missed triggers,
repeated triggers, stuck states, drifting skeleton overlays, delayed
feedback, or anomalies in cameras, models, and page resources after extended
operation. These observations point to a recognition-to-interaction gap
between frame-level recognition outputs and task-level interaction events.

This paper analyzes 8 engineering experiment-and-repair records from a
public-space interactive kiosk project in a scenic-area setting. The
material covers 4 gesture-related interaction tasks: two-hand bowing,
single-hand fist shaking, two-hand catching control, and knowledge-graph
node hovering. From these repair traces, we extract 20 failure instances
and organize them into six working failure classes: model-output
degeneration, temporal mismatch, geometric-scale instability,
coordinate-rendering mismatch, runtime lifecycle failure, and feedback
synchronization and recovery failure.

The repair traces suggest the need for an event-level runtime abstraction
between hand-landmark models and public-space interaction tasks. This
abstraction does not replace the underlying recognition model. Instead, it
involves mechanisms such as temporal confirmation, short-term hand-loss
tolerance, scale normalization, state recovery, coordinate alignment, and
resource lifecycle management, which help organize unstable frame-level
outputs into confirmable or rejectable interaction events. The output of
this paper is not a new gesture-recognition model or a large-scale user
study. It is a working failure taxonomy grounded in repair traces, a
discussion of an event-confirmation runtime abstraction, and a set of
case-study findings.
\end{abstract}


\section{Introduction}\label{sec:intro}

Deploying public-space gesture interaction is challenging not only because
a gesture must be recognized in a single frame, but also because low-level
recognition outputs must be organized into interaction events that users
can perceive. In scenic areas, exhibition spaces, and public service
terminals, vision-based gesture interaction is often used as a low-contact
interaction method. Users do not need to touch a public screen or install
an application. They only need to stand in front of the display and
perform actions such as waving, joining hands, shaking a fist, or hovering
to trigger page feedback.

Field deployment quickly exposes a gap. Gesture recognition that works in
development does not necessarily produce usable interaction events in a
public space. Low-level vision models usually output frame-level hand
landmarks, confidence values, candidate gestures, or hand states. A
public-space interaction system, however, needs user-visible events, such
as completing a two-hand bow, completing a fist-shaking action, moving a
virtual catching object, or hovering over a node in a knowledge graph.
Users do not perceive whether the landmarks in a particular frame are
correct. They perceive whether the event is triggered, whether the state
is stuck, whether the skeleton follows their hand, and whether the prompt
matches the current action.

We call this problem the \emph{recognition-to-interaction gap}. It captures
the situation in which single-frame errors, brief hand losses, coordinate
offsets, state-machine boundary issues, and resource lifecycle problems are
amplified into user-visible interaction failures because the system lacks
event confirmation, runtime state maintenance, and feedback synchronization
mechanisms between recognition outputs and task-level interaction events.

\begin{figure}[t]
  \centering
  \includegraphics[width=\linewidth]{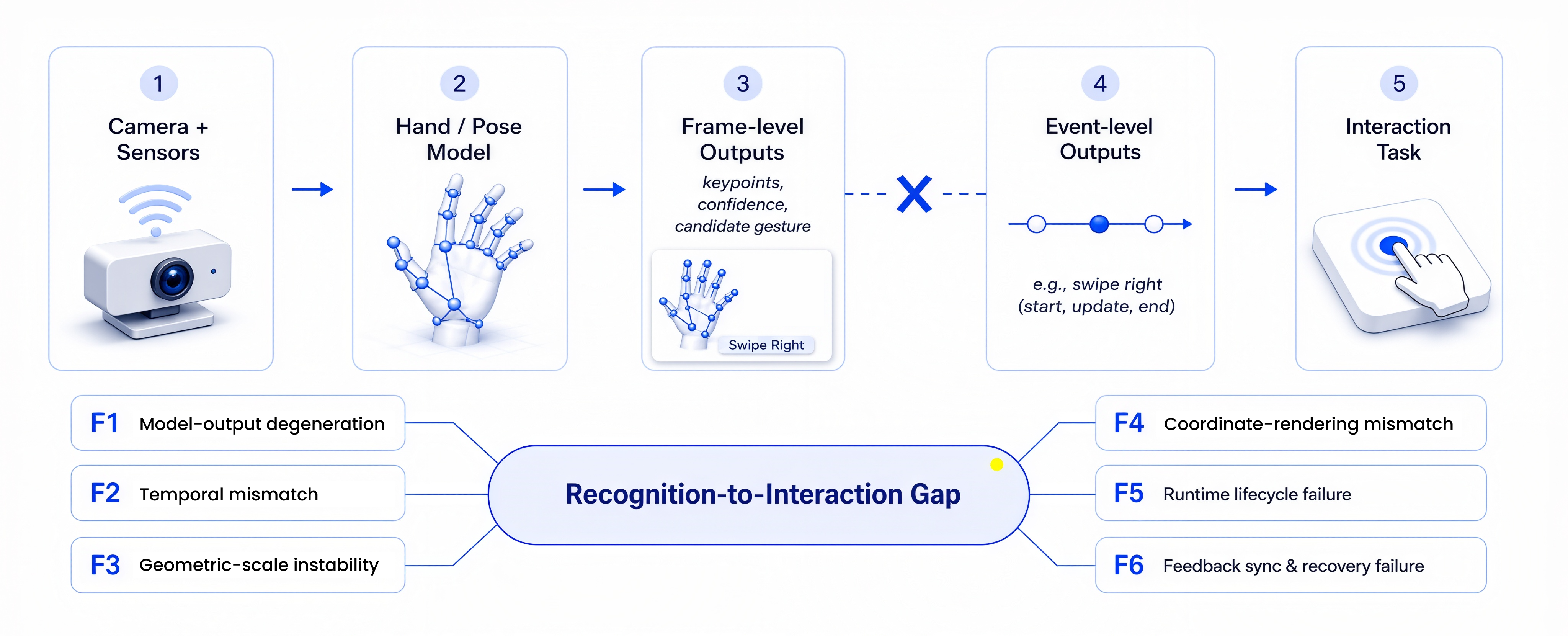}
  \caption{The recognition-to-interaction gap. Frame-level outputs (steps
  1--3) cannot directly become event-level outputs (steps 4--5). Without
  an event-level runtime in between, frame-level errors are amplified into
  six classes of user-visible failures (F1--F6), defined in
  Section~\ref{sec:taxonomy}.}
  \label{fig:gap}
\end{figure}

This gap is more visible in public spaces. Users do not usually follow a
fixed experimental position. Their hands may occupy a small region in the
camera view. The environment may include strong light, backlight, screen
reflections, bystanders, and background interference. The interaction
actions are often continuous processes, such as joining hands and then
bowing, or making a fist and then shaking it. These actions cannot be
handled only by single-frame decisions and require cross-frame state
confirmation. Public-space devices also need to run for extended periods.
The lifecycle of cameras, models, canvases, event listeners, and
\texttt{requestAnimationFrame} (\rAF) callbacks can also affect interaction
reliability.

Rather than improving the underlying hand-recognition model, this paper
examines how recognition outputs are converted into interaction events
during deployment, and how runtime failures emerge in that conversion.

To answer this question, we do not train a new model or conduct a
large-scale user study. Instead, we analyze repair traces from a
public-space interactive kiosk project in a scenic-area setting. The
project contains multiple gesture interaction tasks. The most relevant
tasks for this paper are two-hand bowing, single-hand fist shaking,
two-hand catching control, and knowledge-graph node hovering. During
deployment and debugging, these tasks produced multiple rounds of field
feedback, code diagnosis, parameter changes, state-machine rewrites,
console logs, screenshot observations, resource-release repairs, and
prompt-synchronization repairs.

These materials do not provide a controlled statistical experiment. They
record how failures emerged in field deployment, how repair hypotheses
changed, which repairs appeared to address or fail to address the observed
issues, and which problems recurred across tasks. We therefore treat
repair-trace analysis as a constrained case-study method: we extract
failure instances from field symptoms, engineering hypotheses, code
changes, feedback records, and runtime evidence, map these instances to
failure classes, and use them to locate an event-level runtime layer.

This paper offers three main contributions, all calibrated to the
available material. First, based on 8 engineering experiment-and-repair
records, we extract 20 failure instances and organize six classes of
user-visible failures in public-space gesture interaction. This
classification is not an exhaustive taxonomy of public-space gesture
failures. It is a working taxonomy derived from the current material.
Second, we organize mechanisms that recur in the repair traces, including
short-term hand-loss tolerance, temporal windows, scale normalization,
voting, hysteresis, state recovery, coordinate alignment, and resource
lifecycle management, into an event-level runtime abstraction between
hand-landmark models and interaction tasks. This abstraction is used to
explain failure instances rather than to claim a new algorithm or a
verified performance gain. Third, we summarize five case-study findings
from the 20 failure instances, showing that some field failures cannot be
explained as recognition errors alone but arise from runtime mismatches in
event confirmation, geometric scale, coordinate rendering, lifecycle
management, and feedback synchronization.

The scope is deliberately narrow. The paper is not a new hand-recognition
model or a large-scale user study, and engineering estimates are not
reported as measured empirical results. The goal is to turn public-space
gesture failures from scattered defects and parameter-tuning records into
an analyzable event-confirmation and runtime-failure problem.

\section{Related Work}\label{sec:related}

\subsection{Vision-Based Gesture Recognition}

Vision-based gesture recognition has long studied hand detection, gesture
representation, motion modeling, classification, and human-computer
interaction applications. Pavlovic, Sharma, and Huang provided an early
review of modeling, analysis, and recognition problems in the visual
interpretation of hand gestures~\citep{Pavlovic1997}. Mitra and Acharya
summarized modeling methods, feature representations, and classification
approaches for gesture recognition~\citep{Mitra2007}. Rautaray and Agrawal
surveyed vision-based hand gesture recognition techniques for
human-computer interaction~\citep{Rautaray2015}. In more recent real-time
systems, MediaPipe Hands by Zhang et~al.\ provides an on-device real-time
hand-tracking pipeline from a single RGB camera and is commonly used as a
low-level hand-landmark solution in interaction prototypes~\citep{Zhang2020}.

These studies provide the perceptual basis for this paper, but the focus
here is different. We do not compare the recognition accuracy of
hand-landmark models, nor do we propose a new detection model. We study
how frame-level landmarks, confidence values, and candidate states can
fail to become interaction events in public-space settings after the
underlying model has already produced outputs.

\subsection{Public Displays and Mid-Air Gesture Interaction}

Research on public displays and mid-air gesture interaction examines how
users notice, understand, and use large displays in open spaces. Vogel and
Balakrishnan studied interactive public ambient displays and described
transitions from implicit to explicit interaction and from public to
personal interaction~\citep{Vogel2004}. Hardy, Rukzio, and Davies
observed real-world responses to gesture-based public displays and showed
that public-display gesture interaction is shaped by passers-by,
audiences, distance, discoverability, and social context~\citep{Hardy2011}.
The audience funnel proposed by Michelis and M\"{u}ller also shows that the
process from passing by, noticing, approaching, and explicitly interacting
in public space is unstable~\citep{Michelis2011}.

Public display interaction also involves discoverability, prompting, and
initial-action guidance. The Looking Glass study by M\"{u}ller, Walter,
Bailly, Nischt, and Alt examined how passers-by notice the interactivity
of a shop window~\citep{Muller2012}. StrikeAPose by Walter, Bailly, and
M\"{u}ller studied how to reveal mid-air gestures on public
displays~\citep{Walter2013}. The in-the-wild study by Gentile et~al.\ on
touchless gestural interaction for public displays further shows that
public-space gesture interaction is tied not only to recognition, but also
to user understanding, position, and feedback~\citep{Gentile2015}.

These studies show that public-space gesture interaction cannot be
evaluated only by laboratory recognition accuracy. Prior work on public
displays and mid-air gestures has often focused on discoverability,
interaction stages, and social context. This paper complements that line
of work by focusing on runtime-link failures, including state machines,
coordinate rendering, resource lifecycle, and UI/speech synchronization.

\subsection{Field Deployment, Field Trials, and Repair-Oriented Analysis}

Research on public display deployment emphasizes that real environments
expose user behaviors, environmental conditions, device resources, and
long-running operation issues that are difficult to reproduce in
short-term experiments. The taxonomy of external factors in public display
deployments by M\"{a}kel\"{a} et~al.\ shows that weather, spatial layout,
surroundings, crowd behavior, and site constraints can all affect public
display systems~\citep{Makela2017}.

There is also a tradition in public display and field-trial research of
treating deployed systems and deployment materials as research objects.
Technology probes by Hutchinson et~al.\ provide one example, emphasizing
that systems deployed in real environments can support data collection,
design inspiration, and contextual understanding at the same
time~\citep{Hutchinson2003}. Our use of repair traces is narrower. It
treats engineering repair records, rather than design probes, as the
primary material and does not claim methodological equivalence.

In this paper, repair-trace analysis is used as a constrained case-study
method for engineering repair records. It is not a formal qualitative
coding study and not a controlled user experiment. The method uses failure
symptoms, engineering hypotheses, code changes, and runtime evidence to
derive field failure classes and runtime mechanisms. Wobbrock and Kientz
note that HCI papers can contribute through different forms, including
empirical, artifact, and methodological contributions~\citep{Wobbrock2016}.
In this paper, the contribution lies mainly in a deployment-grounded
case-study taxonomy and runtime abstraction, not in algorithmic
performance or statistical generalization.

\section{Deployment Setting and Evidence Sources}\label{sec:setting}

\subsection{Deployment Setting}

The material comes from a scenic-area public-space interactive kiosk
project. The system runs on large public displays, captures user actions
through a standard camera, and uses hand- or pose-related vision models to
produce frame-level perceptual outputs. Users interact with pages through
gestures such as two-hand bowing, single-hand fist shaking, two-hand
catching control, and knowledge-graph node hovering.

This deployment targets on-site visitors rather than trained experimental
participants. Users do not follow a fixed experimental script and do not
know the limitations of gesture-recognition models. Interaction distance
varies, and at longer distances the hand region becomes much smaller in
the camera view. The device is placed in an open space that may include
strong light, backlight, bystanders, and background interference. It also
needs to run for extended periods. As a result, the management of
resources for cameras, models, canvases, event listeners, and page
switching can all affect the interaction experience.

This paper does not study the cultural content itself, nor does it treat
specific page-level game mechanics as a contribution. These tasks are
used as field public-space interaction tasks for observing how
gesture-recognition outputs are converted into user-visible events.

\subsection{Evidence Sources}\label{sec:evidence}

We compress 8 repair documents into five evidence sources. Table~\ref{tab:evidence}
separates repair-trace evidence, engineering estimates, and future
evaluation designs, so that the existing materials are not misrepresented
as a complete performance experiment.

\begin{table}[ht]
\centering
\caption{Evidence sources from 8 repair documents.}
\label{tab:evidence}
\small
\begin{tabularx}{\linewidth}{@{}p{0.22\linewidth} X X@{}}
\toprule
\textbf{Evidence source} & \textbf{Content} & \textbf{Use in this paper} \\
\midrule
Bowing-module repair log & 6 rounds of repair, \rAF\ callback latency of
226~ms, screenshots, state-machine rewrite & Supports findings on
model-output degeneration, temporal mismatch, and state recovery \\
Fist-shaking repair log & Far-distance fist failures, denominator noise in
scale normalization, EMA and voting repairs & Supports findings on
geometric-scale instability \\
UI and speech-prompt repair log & Step mismatch, speech delay,
completion-prompt overwrite & Supports findings on feedback synchronization \\
Performance assessment report & Static code analysis, cross-code
verification, and expected impact estimates & Used only as engineering
estimates, not empirical results \\
Implementation plan & CPU, GPU, memory, task completion, hand-loss events,
and repeated page-entry metrics & Used as a basis for future small-scale
evaluation \\
\bottomrule
\end{tabularx}
\end{table}

The usable evidence includes field feedback, client feedback, developer
observations, repair rounds, code changes, parameter changes, console
logs, screenshot descriptions, file references, and static code
verification. These forms of evidence can support repair-trace findings,
but they should not be presented as statistical performance experiments.
Some engineering assessment documents contain expected impact estimates,
but this paper does not treat them as empirical results.

\section{Repair-Trace Analysis Method}\label{sec:method}

\subsection{Method Overview}

We use repair-trace analysis as a constrained case-study method rather
than as a controlled user experiment. The goal is to identify analyzable
failure classes and runtime mechanisms from field deployment repairs. The
analysis process consists of six steps: field symptom, engineering
hypothesis, code or parameter change, feedback record or runtime evidence,
failure class, and runtime mechanism.

This method fits the material because the documents record not only final
repairs but also failed attempts, feedback records, console warnings,
rollbacks, and changes in repair hypotheses. For example, the two-hand
bowing task was not fixed by a single parameter adjustment. The repair
process went through throttling, tolerance adjustment, confidence
adjustment, input-resolution reduction, one-hand maintenance,
inference-backend checks, and task-rule correction before converging on a
new state machine. This process shows that the problem is not only an
unsuitable threshold. It also reflects a missing semantic recovery path
between degraded low-level outputs and the task state machine.

\begin{figure}[t]
  \centering
  \includegraphics[width=0.85\linewidth]{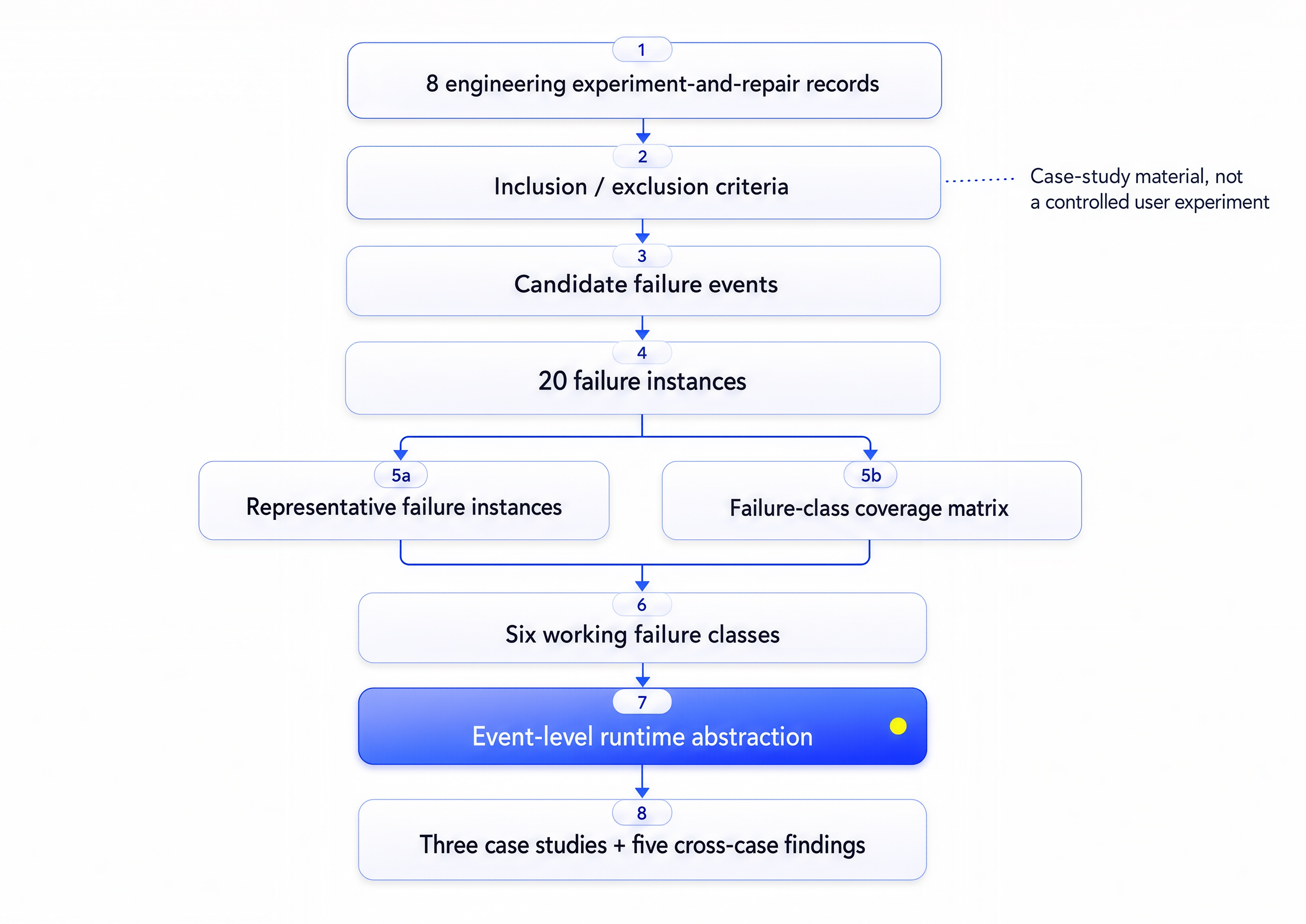}
  \caption{Repair-trace analysis pipeline. Engineering repair records are
  compressed into failure instances, a working taxonomy, and an
  event-level runtime abstraction. The dashed annotation reminds the
  reader that the inputs are case-study material, not data from a
  controlled user experiment.}
  \label{fig:pipeline}
\end{figure}

\subsection{Inclusion and Exclusion Criteria}

The initial extraction of failure instances was conducted by the system
developer who maintained the relevant gesture modules, and the extracted
instances were reviewed against representative runtime evidence. This
paper does not present the process as a formal qualitative coding study
and does not claim inter-coder agreement among multiple independent coders.

To reduce retrospective interpretation bias, we include only failure
instances that satisfy all of the following conditions. First, the issue
must have a user-visible symptom or a runtime-visible anomaly. Second, the
issue must have at least one traceable form of evidence, such as field
feedback, client feedback, developer observation, console logs,
screenshot descriptions, code changes, commit records, or static code
verification. Third, the issue must correspond to at least one engineering
hypothesis, repair attempt, or runtime mechanism. Fourth, the issue must
be representable as a chain linking symptom, condition, source, repair,
and failure class.

We exclude records that contain only parameter suggestions without symptom
or runtime evidence, UI text changes that do not affect gesture-event
state or user understanding of the event, expected impact estimates
without implemented repairs, and speculative issues that cannot be traced
to a specific task, module, log, screenshot, or code change.

The 8 documents selected for this paper cover the main repair records
directly related to gesture recognition, event triggering, prompt
synchronization, and runtime state maintenance in the project. They do
not include documents that only concern UI styling, content copy, or
business-process changes. After scanning these repair records, we
deduplicate candidate failure events and organize them into 20 failure
instances. The current sample boundary includes 8 engineering
experiment-and-repair records, 4 gesture-related interaction tasks, 6
recognition iterations for two-hand bowing, 3 recognition iterations for
single-hand fist shaking, 7 specific defects or design decisions in
prompt synchronization, and multiple implemented repairs and static checks
covered in the post-repair performance assessment.

Accordingly, the material is used for case-study interpretation rather
than for claims about large user samples, statistical significance,
measured accuracy changes, or user recovery rates.

\subsection{Evidence-Type Labels}

For readability, we group evidence into five labels. A failure instance
may have more than one evidence type.

\begin{table}[ht]
\centering
\caption{Evidence-type labels used in failure instance tables.}
\label{tab:evidence-labels}
\small
\begin{tabularx}{0.85\linewidth}{@{}lX@{}}
\toprule
\textbf{Evidence type} & \textbf{Meaning} \\
\midrule
Field feedback     & Feedback from field use, clients, or developer observation on site \\
Runtime log        & Console records, runtime warnings, or performance logs \\
Code change        & Code changes, parameter changes, or state-machine rewrites \\
Static verification & Cross-code verification, static checking, or post-repair code review \\
Visual trace       & Screenshots, video-frame descriptions, or visible anomaly records \\
\bottomrule
\end{tabularx}
\end{table}

\subsection{Representative Failure Instances}

The main text lists only representative failure instances that support the
main findings. To avoid confusion with the 20 underlying failure
instances, Table~\ref{tab:representative} includes a column that maps each
representative instance to the underlying instance or instances. The full
list of 20 failure instances is provided in Appendix~\ref{app:fi-table}.

\begin{table}[ht]
\centering
\caption{Representative failure instances supporting the main findings.}
\label{tab:representative}
\footnotesize
\begin{tabularx}{\linewidth}{@{}llp{0.10\linewidth}p{0.16\linewidth}p{0.16\linewidth}p{0.16\linewidth}X@{}}
\toprule
\textbf{R} & \textbf{FI} & \textbf{Task} & \textbf{Symptom} & \textbf{Runtime source} & \textbf{Repair action} & \textbf{Evidence} \\
\midrule
R1 & FI-02 & Two-hand bowing & Fast bowing is missed
   & \rAF\ callback latency is 226~ms; effective frame rate is low
   & Reduce input resolution; reset state-machine temporal window
   & Runtime log; field feedback; code change \\
R2 & FI-04 & Two-hand bowing & State freezes after one-hand degeneration
   & \texttt{length < 2 return} prevents the vertical midpoint from updating
   & One-hand maintenance, hysteresis exit, and state-machine rewrite
   & Code change; static verification \\
R3 & FI-06, FI-07, FI-08 & Fist shaking & Far-distance fist is lost or flips briefly
   & Denominator noise in scale normalization; single-point flipping rule
   & Multi-vector palm scale; EMA; voting
   & Field feedback; code change; static verification \\
R4 & FI-09 & Fist shaking & Skeleton does not follow the hand during large-amplitude shaking
   & Landmark drawing lacks smoothing; tracking confidence unstable near boundary
   & Drawing smoothing; tracking-threshold adjustment
   & Field feedback; visual trace; code change \\
R5 & FI-18 & UI and speech prompt & Completion prompt is not visible
   & Audio-failure branch triggers the completion callback synchronously
   & Minimum display time; step guard
   & Field feedback; code change \\
\bottomrule
\end{tabularx}
\end{table}

\subsection{Failure-Class Coverage}\label{sec:coverage}

This section reports how the six failure classes are covered in the
current repair material. Counts indicate coverage in repair traces, not
statistical frequency in field deployment.

\begin{table}[ht]
\centering
\caption{Failure-class coverage matrix across interaction tasks. Counts
are repair-trace coverage, not deployment frequency.}
\label{tab:coverage}
\small
\begin{tabular}{@{}lcccccc@{}}
\toprule
\textbf{Failure class} & \textbf{Bowing} & \textbf{Fist} & \textbf{Catching} & \textbf{Hovering} & \textbf{UI/Speech} & \textbf{Total} \\
\midrule
F1\ Model-output degeneration                 & 2 & 0 & 0 & 0 & 0 & 2 \\
F2\ Temporal mismatch                          & 4 & 0 & 1 & 1 & 0 & 6 \\
F3\ Geometric-scale instability                & 0 & 3 & 1 & 0 & 0 & 4 \\
F4\ Coordinate-rendering mismatch              & 0 & 1 & 2 & 1 & 0 & 4 \\
F5\ Runtime lifecycle failure                  & 0 & 1 & 0 & 1 & 1 & 3 \\
F6\ Feedback synchronization and recovery failure & 0 & 0 & 0 & 0 & 3 & 3 \\
\bottomrule
\end{tabular}
\end{table}

\begin{figure}[t]
  \centering
  \includegraphics[width=0.95\linewidth]{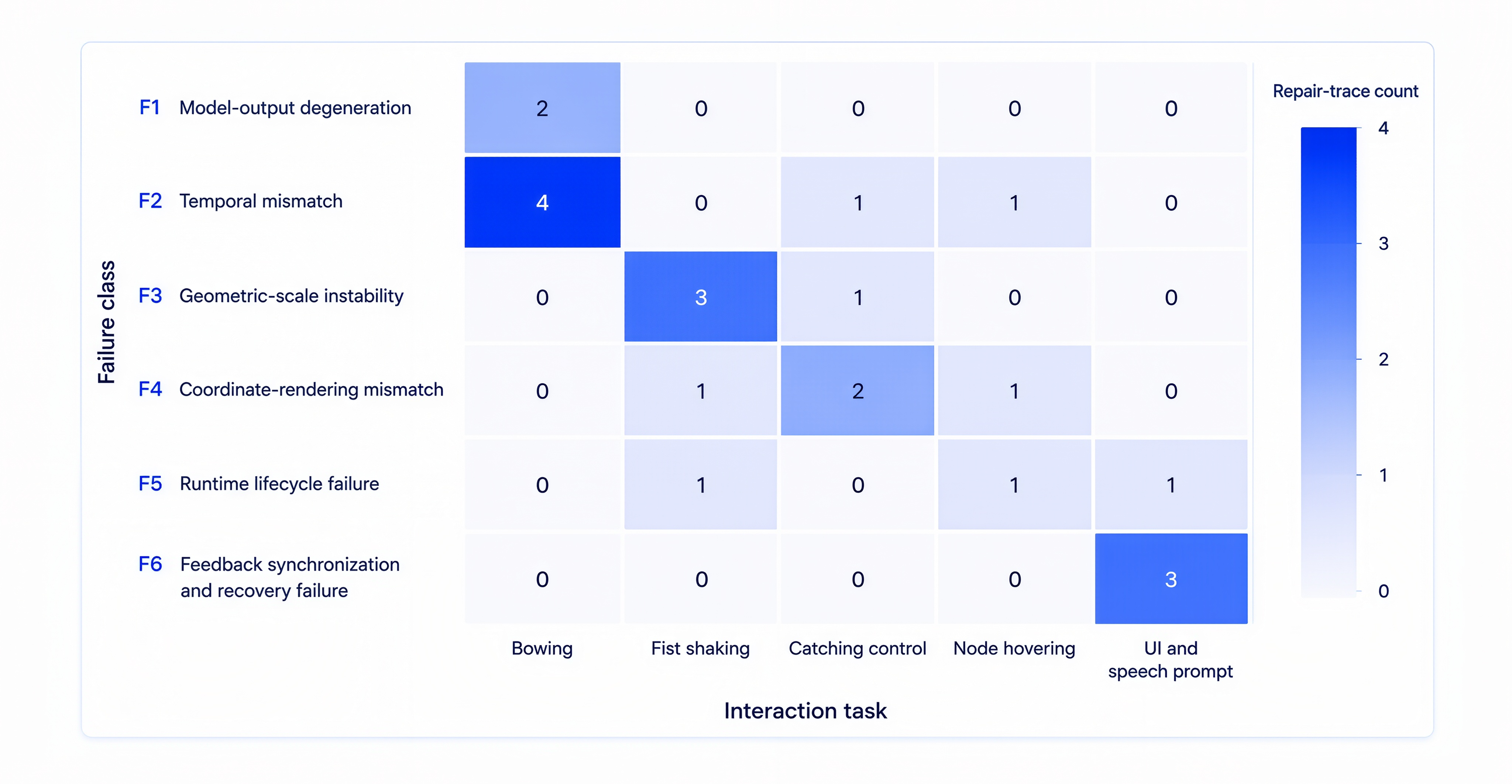}
  \caption{Visualization of the failure-class coverage matrix
  (Table~\ref{tab:coverage}). Color intensity reflects per-cell repair-trace
  count. The matrix indicates coverage in the present material, not
  statistical frequency in deployment.}
  \label{fig:coverage}
\end{figure}

Some underlying instances are counted in more than one failure class, so
the class totals exceed 20. The multi-class instances are FI-04, which is
counted in both F1 and F2, and FI-12, which is counted in both F3 and F4.
F1 and F6 have relatively narrow coverage in the current material,
reflecting the limited coverage of overlapping two-hand tasks and
feedback-synchronization issues. This limitation is discussed further in
Section~\ref{sec:limitations}.

\subsection{Validity Threats of Repair-Trace Analysis}\label{sec:threats}

This constrained case-study method has several validity threats. First,
the 20 failure instances are not random samples. They come from
deployment repair records in one project. They can characterize failure
patterns in this case, but they cannot be used to infer the statistical
distribution of public-space gesture-interaction failures. Second, the
instances were extracted by the system developer, which introduces
developer-as-analyst bias. We reduce this bias in three ways: each
failure instance must have at least one traceable form of evidence;
records with only parameter suggestions or expected impact estimates are
excluded; and representative instances are mapped to underlying instance
identifiers. These measures reduce but do not remove bias. Third, the
current material is more likely to preserve issues that were located or
repaired, while unrecorded, unreproduced, or abandoned issues may be
underrepresented. Fourth, the six failure classes form a working taxonomy
rather than an exhaustive taxonomy. Fifth, the paper does not include
external coder review, user experiments, or ablation studies, so the
mechanisms are not treated as sources of quantified effect improvement.

\section{Field Failure Classification}\label{sec:taxonomy}

The six failure classes proposed in this paper are not an exhaustive
classification of public-space gesture-interaction failures. They are a
working classification derived from the current 8 repair records and 20
failure instances. Section~\ref{sec:coverage} describes how these classes
are covered by the material. This section defines the classes. Multi-class
instances such as FI-04 (F1 and F2) and FI-12 (F3 and F4) appear under
their primary class in this table; full class membership is reported in
Section~\ref{sec:coverage} and Appendix~\ref{app:fi-table}.

\begin{table}[ht]
\centering
\caption{Six working failure classes for public-space gesture interaction.}
\label{tab:classes}
\footnotesize
\begin{tabularx}{\linewidth}{@{}lp{0.18\linewidth}Xp{0.16\linewidth}X@{}}
\toprule
\textbf{ID} & \textbf{Name} & \textbf{Definition} & \textbf{Representative FI} & \textbf{User-visible consequence} \\
\midrule
F1 & Model-output degeneration
   & The user action is semantically valid, but the low-level model output
     degenerates into incomplete or unstable frame-level results
   & FI-01, FI-04
   & Stuck action; missed trigger; state not advancing \\
F2 & Temporal mismatch
   & Model inference rhythm, rendering rhythm, and action duration do not
     match, preventing event confirmation
   & FI-02, FI-03, FI-05, FI-11, FI-14
   & Event cannot form; fast action is missed \\
F3 & Geometric-scale instability
   & Small hand scale or landmark jitter is amplified by ratio computation,
     causing state flips
   & FI-06, FI-07, FI-08, FI-12
   & Fist-state flipping; far-distance misclassification; jitter \\
F4 & Coordinate-rendering mismatch
   & Model coordinates, video cropping, canvas drawing, and screen layout
     are inconsistent
   & FI-09, FI-13, FI-15
   & Skeleton or interaction object appears offset \\
F5 & Runtime lifecycle failure
   & Camera, model, listener, canvas, or render loop is not released or
     recovered correctly
   & FI-10, FI-16, FI-17
   & Stutter, leakage, or abnormal behavior after long operation \\
F6 & Feedback synchronization and recovery failure
   & Visual recognition, task state machine, UI text, and speech prompt
     are not synchronized
   & FI-18, FI-19, FI-20
   & Users cannot understand the current state or how to continue \\
\bottomrule
\end{tabularx}
\end{table}

\section{Event-Level Runtime Abstraction}\label{sec:runtime}

\subsection{Runtime Position}

Based on the failure classification above, we abstract the observed repair
mechanisms into an event-level runtime layer. This layer is not a new
hand-recognition model and not a complete platform. It sits between the
low-level hand or pose model and the public-space interaction task. Its
goal is not to make every frame correct, but to confirm user intent from
unstable frame-level outputs and prevent short-term errors from becoming
user-visible failures. Within this layer, we organize mechanisms into
three sublayers: event confirmation, runtime maintenance, and feedback
synchronization.

\begin{figure}[t]
  \centering
  \includegraphics[width=\linewidth]{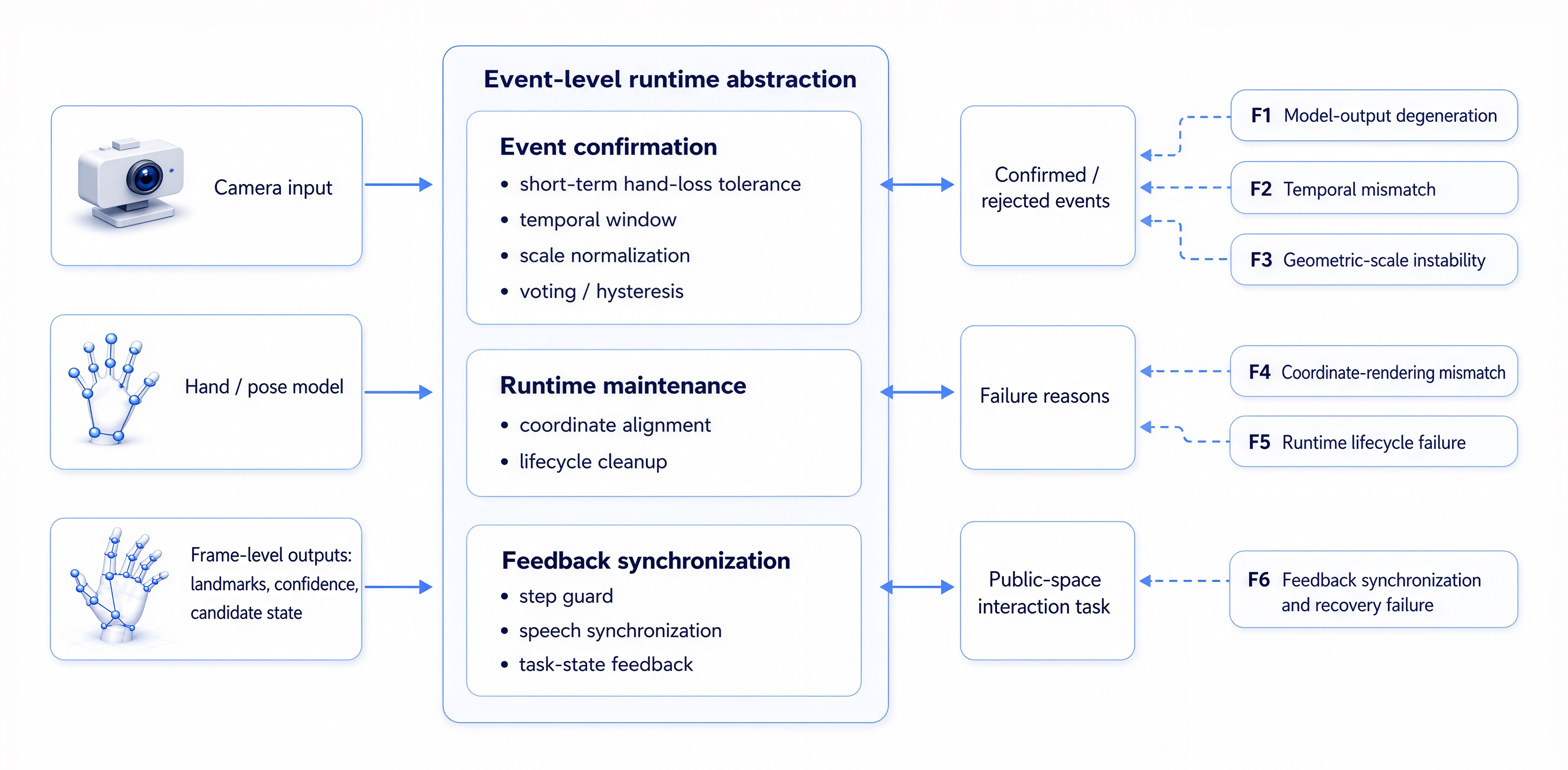}
  \caption{Event-level runtime abstraction. A runtime layer organizes
  unstable frame-level outputs into confirmable or rejectable interaction
  events. Three sublayers (event confirmation, runtime maintenance, and
  feedback synchronization) address different failure classes (F1--F6)
  defined in Section~\ref{sec:taxonomy}.}
  \label{fig:runtime}
\end{figure}

\subsection{Event-Level Confirmation Procedure}\label{sec:procedure}

The event-level confirmation procedure receives frame-level hand
landmarks, confidence values, candidate gestures, and timestamps, while
maintaining the current task phase, recent valid frames, and cooldown
state. It first checks whether the observation falls inside the valid
interaction region and rejects low-confidence or physically implausible
observations. It then performs geometric normalization using hand scale or
task-related body scale and smooths short-term motion or scale estimates
when needed. If the hand is briefly lost, the system retains the previous
semantic state within a tolerance window instead of resetting immediately.
Candidate gestures are aggregated within a temporal window, and voting or
hysteresis is used to reduce boundary flips. The candidate is then checked
against the legal transitions in the current task phase. The procedure
outputs a confirmed event only when the confirmation signal reaches the
task threshold and the system is not in cooldown. Otherwise, it rejects
the candidate and records the corresponding failure reason.

To avoid presenting engineering heuristics as an evaluated algorithm, this
paper does not provide a concrete scoring formula in the main text and
does not report any optimal combination of weights or thresholds. The
following expression is only used to describe the components of
event-confirmation signals. It is not used in any experiment in this paper
and is not a methodological contribution. Conceptually, event confirmation
can be represented as a task-related decision function:

\begin{align}
  s_t       &= g_k(c_t,\, q_t,\, r_t,\, h_t,\, p_t), \label{eq:score} \\
  \hat{e}_t &= \mathbb{I}\!\left(s_t \geq \tau_k \,\land\, \neg b_t\right). \label{eq:event}
\end{align}

Here, $t$ denotes the current frame, $k$ denotes the task type, $c_t$
denotes confidence information within the temporal window, $q_t$ denotes
temporal consistency, $r_t$ denotes scale validity, $h_t$ denotes
state-transition validity, $p_t$ denotes penalty terms such as hand loss,
candidate switching, and cooldown, $b_t$ denotes whether the system is
currently in cooldown, $\tau_k$ is a task-related threshold, and
$\mathbb{I}(\cdot)$ is the indicator function. The function $g_k(\cdot)$
only represents a task-related heuristic combination and should not be
read as a portable numerical model fitted by this paper.

\subsection{Mapping Runtime Mechanisms to Failure Classes}

\begin{table}[ht]
\centering
\caption{Mapping between runtime mechanisms, sublayers, and failure
classes (Section~\ref{sec:taxonomy}).}
\label{tab:mapping}
\small
\begin{tabularx}{\linewidth}{@{}p{0.27\linewidth}p{0.20\linewidth}p{0.10\linewidth}X@{}}
\toprule
\textbf{Runtime mechanism} & \textbf{Sublayer} & \textbf{Class} & \textbf{Supporting material} \\
\midrule
Short-term hand-loss tolerance     & Event confirmation       & F1, F2 & Bowing module; fist-shaking module \\
Temporal window and state machine  & Event confirmation       & F2     & Bowing module \\
Scale normalization                & Event confirmation       & F3     & Fist-shaking module \\
EMA and voting                     & Event confirmation       & F3, F4 & Fist-shaking module \\
Coordinate alignment               & Runtime maintenance      & F4     & Catching control; node hovering \\
Lifecycle cleanup                  & Runtime maintenance      & F5     & Performance assessment; implementation records \\
Step guard and speech synchronization & Feedback synchronization & F6  & UI and speech-prompt repair log \\
\bottomrule
\end{tabularx}
\end{table}

\section{Repair-Trace Evidence and Case-Study Findings}\label{sec:cases}

\subsection{Bowing Module: Parameter Tuning Does Not Replace State Recovery}\label{sec:bowing}

The bowing-module repair trace provides a complete chain of field
feedback, runtime logs, visual observations, code diagnosis, and repair
actions. This chain is a representative repair trace, not a performance
experiment.

\begin{figure}[t]
  \centering
  \includegraphics[width=\linewidth]{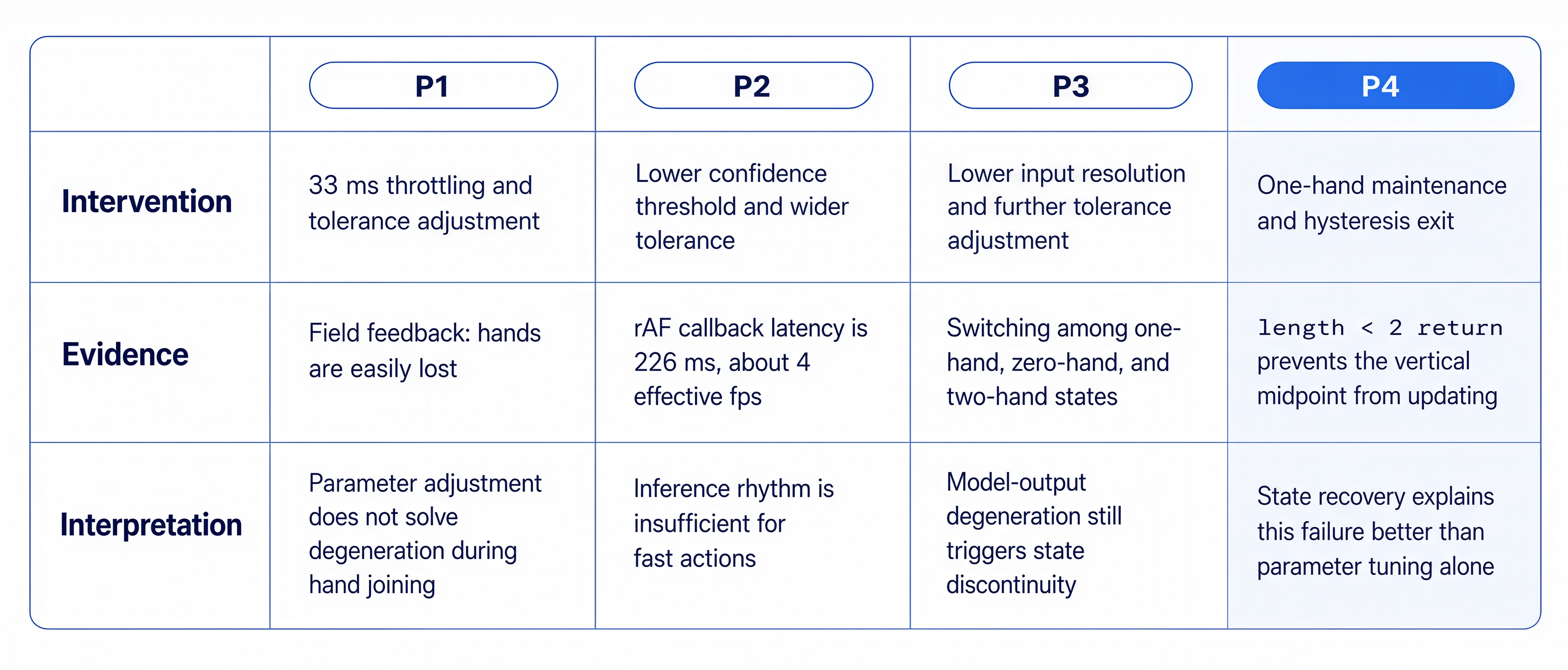}
  \caption{Representative repair trace of the two-hand bowing task. Three
  rounds of parameter tuning (P1--P3) failed to resolve the failure;
  state recovery (P4) explained what tuning could not.}
  \label{fig:bowing-trace}
\end{figure}


This case shows that multiple user-visible failures cannot be explained as
low-level recognition errors alone. FI-01 to FI-05 show that some failures
require tracing model-output degeneration, state-machine boundaries, and
temporal windows together. This does not imply that state recovery
statistically improves task completion. It only shows that, in this case,
threshold adjustment alone does not explain or handle all observed
failures.

\subsection{Fist-Shaking Module: Far-Distance Fist Failure and Scale-Sensitive Noise}

The far-distance fist-shaking issue also forms a traceable chain. User
feedback reports that the fist is not detected or is detected and then
quickly lost. Code diagnosis points to denominator noise in scale
normalization, far-distance landmark noise, and a single-point flipping
rule. The repair then shifts to multi-vector scale estimation, EMA, and
voting.

\begin{table}[ht]
\centering
\caption{Fist-shaking module repairs.}
\label{tab:fist}
\footnotesize
\begin{tabularx}{\linewidth}{@{}p{0.22\linewidth}p{0.22\linewidth}Xp{0.18\linewidth}@{}}
\toprule
\textbf{Problem} & \textbf{Before repair} & \textbf{Repair action} & \textbf{Evidence level} \\
\midrule
Denominator noise in scale normalization
   & Single vector from wrist to landmark 9
   & Average of vectors from wrist to landmarks 5, 9, 13, and 17
   & Code change; static verification \\
Ratio jitter
   & No temporal smoothing
   & Palm-scale EMA with smoothing coefficient 0.2
   & Code change \\
Single-point flip
   & Any fingertip crossing the threshold marks the hand as non-fist
   & At least 3 fingertips must satisfy the condition to confirm a fist
   & Code change; static verification \\
Far-distance fist loss
   & Confidence threshold 0.3 and tolerance window 15 frames
   & Confidence threshold 0.2 and tolerance window 30 frames
   & Code change; field feedback \\
\bottomrule
\end{tabularx}
\end{table}

This case supports the finding that far-distance fist failures can be
explained as a runtime problem involving geometric scale, temporal
smoothing, and redundant decision rules, rather than as a simple
confidence-threshold problem. The material does not include repeated
trials, an annotation scheme, or statistical testing, so we do not report
an accuracy result.

\subsection{UI and Speech Prompts: Feedback Synchronization Shapes the Perceived Success of Events}

The UI and speech-prompt repair records show that interaction failures can
occur after an event has been or is about to be confirmed. Even when a
gesture event is available, inconsistent UI text and speech state can
still make the user perceive a failure.

\begin{table}[ht]
\centering
\caption{UI and speech-prompt repairs.}
\label{tab:ui-speech}
\footnotesize
\begin{tabularx}{\linewidth}{@{}p{0.24\linewidth}p{0.24\linewidth}Xp{0.20\linewidth}@{}}
\toprule
\textbf{Failure symptom} & \textbf{Runtime source} & \textbf{Repair action} & \textbf{Role in this paper} \\
\midrule
Completion prompt is not visible
   & Audio-failure branch advances the step synchronously
   & Minimum display time and step guard
   & Supports feedback synchronization failure \\
Text does not roll back after bowing timeout
   & Steps 5, 8, and 11 lack fallback handling for the joined-hands state
   & Roll back to the previous prompt under the joined-hands state
   & Supports feedback recovery failure \\
Speech prompt lags behind
   & 800~ms debounce overlaps with remaining speech playback time
   & Shorter debounce and fast path for core actions
   & Supports mismatch between event state and prompt state \\
\bottomrule
\end{tabularx}
\end{table}

This material supports the observation that successful public-space
gesture interaction depends not only on detection and event confirmation,
but also on whether UI and speech prompts are synchronized with the event
state. It does not support a quantified claim about user-experience
improvement.

\subsection{Cross-Case Findings}

\paragraph{Finding 1.} Some user-visible failures cannot be explained by
low-level recognition errors alone. This finding is supported by
representative instances R1 and R2 and underlying instances FI-01 to
FI-05. The bowing repair trace shows that fast bowing misses, two-hand
degeneration into one hand, and state-machine freezing are not problems
at the same level. Together, they point to mismatches among frame-level
output degeneration, temporal windows, and task state machines.

\paragraph{Finding 2.} Far-distance fist failure can be explained as
scale-sensitive geometric noise. This finding is supported by
representative instance R3 and underlying instances FI-06 to FI-08. At
longer distances, the hand occupies a smaller image region, and small
landmark jitter is amplified through ratio computations. If the decision
rule also depends on a single-point threshold, the system can flip
quickly between fist and non-fist states.

\paragraph{Finding 3.} Rendering alignment affects user-visible
interaction correctness. This finding is supported by representative
instance R4 and underlying instances FI-09, FI-13, and FI-15. In
public-display interaction, users judge whether the system understands
their action by looking at skeleton overlays, halos, or interaction
objects on the screen. Even when the underlying landmarks are usable,
inconsistent coordinate mapping, video cropping, and canvas drawing can
produce visible offsets.

\paragraph{Finding 4.} Lifecycle cleanup repeatedly appears in repairs
related to long-running deployment. This finding is supported by
underlying instances FI-10, FI-16, and FI-17. A public-space kiosk differs
from a one-time demo. It needs to run for extended periods and repeatedly
handle users entering and leaving pages. The release and recovery of
cameras, models, listeners, canvases, and render loops affect whether the
system remains usable.

\paragraph{Finding 5.} Feedback synchronization and fast-motion failures
expose risks beyond event confirmation. This finding is supported by
representative instance R5, underlying instances FI-18 to FI-20, and the
negative cases FI-02 and FI-05. Overwritten completion prompts, missing
timeout rollback, and delayed speech prompts show that user-visible
failures can still occur after event confirmation. At the same time, fast
bowing, inference latency, and state-transition conditions remain in
tension. Future claims about improved stability would require adaptive
inference and rendering schedules, as well as before-after or ablation
evidence.

\section{Discussion, Limitations, and Conclusion}\label{sec:discussion}

\subsection{Discussion}

This paper reframes public-space gesture interaction from a frame-level
recognition problem to an event-level confirmation problem. This
reframing matters because user-visible failures are not equivalent to
errors in a single model output frame. A system may produce reasonable
landmarks in most frames and still fail because of state machines,
temporal windows, geometric scale, coordinate mapping, resource release,
or prompt synchronization.

The event-level runtime abstraction proposed in this paper does not depend
on the novelty of each individual mechanism. Debouncing, EMA, voting,
hysteresis, state machines, and resource release are common engineering
mechanisms. The main point is that, in public-space gesture interaction,
these mechanisms should not remain scattered patches. They should be
organized as a runtime layer between model outputs and interaction tasks,
so that user-visible failures can be analyzed.

Repair traces are useful because they preserve how failure hypotheses
change. For example, the bowing repair trace shows that the initial
hypothesis focused on inference latency, tolerance windows, and
confidence thresholds, but later diagnosis pointed to the inability of the
state machine to handle degraded outputs. The fist-shaking repair trace
shows that far-distance detection failure is not solved simply by lowering
the confidence threshold. It results from denominator noise in scale
normalization, single-point flipping, and insufficient temporal redundancy.

\subsection{Limitations}\label{sec:limitations}

This paper has several limitations. First, it is not a controlled user
experiment and does not include a large user sample. The material comes
from field-deployment repair processes and supports case-study findings
rather than statistically significant performance evaluation. Second, some
values in the post-repair performance assessment come from static code
analysis, cross-code verification, and engineering estimates, and should
not be interpreted as measured empirical results. Third, the failure
instances were extracted by the system developer and checked against
representative runtime evidence. This is not a formal qualitative coding
study. Fourth, the six failure classes form a working taxonomy based on
the current material and do not constitute a complete taxonomy of
public-space gesture-interaction failures. Fifth, F1 in the current
dataset mainly comes from overlapping two-hand gesture tasks, and F6
mainly comes from UI and speech-prompt repairs. Cross-task transfer still
needs further study. Sixth, some state-machine rewrites require stricter
field validation. Seventh, failure recovery is currently represented only
by prompt synchronization and diagnostic plans. There are no data on
recovery rate, mean recovery time, or user abandonment rate, so recovery
cannot be treated as a measured main contribution. Eighth, primary-user
locking in multi-person scenes, child users, night lighting, and
cross-device transfer have not been systematically covered.

\subsection{Conclusion}

This paper studies the recognition-to-interaction gap in public-space
gesture interaction through 8 engineering experiment-and-repair records
from 4 gesture interaction tasks in a scenic-area public-space interactive
kiosk. Using repair-trace analysis as a constrained case-study method, we
extract 20 failure instances and organize field failures into six working
failure classes: model-output degeneration, temporal mismatch,
geometric-scale instability, coordinate-rendering mismatch, runtime
lifecycle failure, and feedback synchronization and recovery failure.

Based on these observations, the paper develops an event-level runtime
abstraction that organizes short-term hand-loss tolerance, temporal
confirmation, scale normalization, voting, hysteresis, state recovery,
coordinate alignment, and resource lifecycle management between
frame-level outputs and interaction events. Rather than proposing a new
gesture-recognition model or a large-scale empirical study, it shows how
repair traces can reveal runtime problems between frame-level recognition
and event-level confirmation in public-space gesture interaction.

\bibliographystyle{plainnat}
\bibliography{refs}

\appendix

\section{Full Failure Instance Table}\label{app:fi-table}

\begin{small}
\begingroup
\setlength{\tabcolsep}{3pt}
\renewcommand{\arraystretch}{0.95}
\begin{longtable}{@{}P{0.05\linewidth}P{0.13\linewidth}P{0.13\linewidth}P{0.10\linewidth}P{0.13\linewidth}P{0.13\linewidth}P{0.12\linewidth}P{0.06\linewidth}@{}}
\caption{Full list of 20 failure instances.}\label{tab:fi-full}\\
\toprule
\textbf{ID} & \textbf{Module/Task} & \textbf{Symptom} & \textbf{Trigger}
& \textbf{Runtime source} & \textbf{Repair action} & \textbf{Evidence}
& \textbf{Class} \\
\midrule
\endfirsthead
\multicolumn{8}{l}{\small\itshape (Table~\ref{tab:fi-full} continued)} \\
\toprule
\textbf{ID} & \textbf{Module/Task} & \textbf{Symptom} & \textbf{Trigger}
& \textbf{Runtime source} & \textbf{Repair action} & \textbf{Evidence}
& \textbf{Class} \\
\midrule
\endhead
\bottomrule
\endfoot

FI-01 & Two-hand bowing & Joined hands recognized as one hand or lost
      & Overlapping joined hands & Model output degenerates from two hands
        to one or zero hands
      & One-hand maintenance, hysteresis exit, state-machine rewrite
      & Field feedback; visual trace; code change & F1 \\
FI-02 & Two-hand bowing & Fast bowing not recognized
      & Fast action; high inference latency
      & \rAF\ callback latency 226~ms; few frames capture the action
      & Lower input resolution; state-machine temporal window
      & Runtime log; field feedback; code change & F2 \\
FI-03 & Two-hand bowing & Hands more easily lost after throttling
      & 33~ms throttling during joined-hands action
      & Throttling changes physical duration of hand-loss tolerance
      & Adjust hand-loss tolerance window
      & Field feedback; code change & F2 \\
FI-04 & Two-hand bowing & State machine freezes after \texttt{length < 2}
      & Two hands degenerate into one hand
      & Code returns immediately; lacks semantic recovery
      & Four-stage state machine
      & Code change; static verification & F1, F2 \\
FI-05 & Two-hand bowing & Bowing phase stuck or release condition unmet
      & User does not raise head after bowing; motion amplitude differs
      & Return condition too strict; direction rule incorrect
      & Task-rule correction; state-machine rewrite
      & Code change; static verification & F2 \\
FI-06 & Single-hand fist shaking & Far-distance fist not detected
      & Distance 2.0--3.0~m
      & Hand scale too small; confidence and geometric ratios unstable
      & Lower confidence threshold; extend tolerance window
      & Field feedback; code change & F3 \\
FI-07 & Single-hand fist shaking & Fist briefly detected then lost
      & Palm scale jitters near threshold
      & Single-vector denominator noise from wrist to landmark 9
      & Average four palm-base vectors
      & Static verification; code change & F3 \\
FI-08 & Single-hand fist shaking & Single-fingertip jitter causes non-fist
      & Occasional far-distance landmark jump
      & Any fingertip crossing threshold flips state
      & Voting rule requiring at least three valid fingertips
      & Code change; static verification & F3 \\
FI-09 & Single-hand fist shaking & Skeleton does not follow hand during shaking
      & Fast large-amplitude shaking
      & Inference throttling; tracking delay; unsmoothed drawing
      & Drawing smoothing; tracking-confidence callback
      & Field feedback; visual trace; code change & F4 \\
FI-10 & Single-hand fist shaking & State residue across users after no-hand
      & User leaves and another enters
      & Peak-valley detection state not reset
      & Reset peak-valley state when no hands detected
      & Code change; static verification & F5 \\
FI-11 & Two-hand catching control & Single-frame miss causes basket flicker
      & Fast horizontal movement; transient occlusion
      & No-hand state reported immediately; no debounce
      & Three-frame no-hand debounce
      & Code change; static verification & F2 \\
FI-12 & Two-hand catching control & Basket or skeleton high-frequency jitter
      & Per-frame x-coordinate directly drives interaction
      & Hand position lacks smoothing
      & EMA smoothing on x-coordinate
      & Visual trace; code change & F3, F4 \\
FI-13 & Two-hand catching control & Skeleton offset from hand
      & Video style transform and canvas coordinates inconsistent
      & Coordinate mapping and rendering path separated
      & Unified rendering path; shared coordinate transform
      & Visual trace; code change & F4 \\
FI-14 & Knowledge-graph node hovering & Skeleton lost during fast shaking
      & Fast movement; transient no-hand state
      & No-hand debounce and inference throttling missing
      & No-hand debounce; 33~ms throttling
      & Visual trace; code change & F2 \\
FI-15 & Knowledge-graph node hovering & Skeleton offset from hand position
      & Video crop and canvas coordinates inconsistent
      & Style rendering and script coordinates cannot fully align
      & Dual canvas; unified drawing; shared crop parameters
      & Visual trace; code change & F4 \\
FI-16 & Knowledge-graph node hovering & Long-running operation has resource-leak risk
      & Repeated page entry/exit; long operation
      & Stop routine does not release model, listener, or video stream
      & Full release of hand-overlay resources
      & Code change; static verification & F5 \\
FI-17 & Model or camera runtime & Resource leak after init failure
      & Parallel initialization failure
      & One resource not released when other branch fails
      & Mutual release after parallel initialization results
      & Code change; static verification & F5 \\
FI-18 & UI and speech prompt & ``The N-th incense has been offered'' not shown
      & Audio failure or browser autoplay blocking
      & Audio-failure branch calls completion callback synchronously and
        overwrites prompt
      & Minimum display time; step guard
      & Field feedback; code change & F6 \\
FI-19 & UI and speech prompt & Text does not roll back after bowing timeout
      & Bowing depth not reached; phase returns to ready
      & Some steps lack fallback handling for joined-hands state
      & Add fallback for joined-hands state
      & Field feedback; code change & F6 \\
FI-20 & UI and speech prompt & Fortune-shaking prompt lags behind
      & Gesture state changes quickly while speech is playing
      & 800~ms debounce overlaps with remaining speech playback time
      & Shorter debounce; fast path for core actions
      & Field feedback; runtime log; code change & F6 \\
\end{longtable}
\endgroup
\end{small}

\section{Repair Document List}\label{app:doc-list}

\begin{table}[ht]
\centering
\caption{Eight repair documents from which the failure instances and case
studies were extracted.}
\label{tab:docs}
\small
\begin{tabularx}{\linewidth}{@{}llX@{}}
\toprule
\textbf{ID} & \textbf{Document} & \textbf{Source file} \\
\midrule
D1 & PrayerTracker experiment record, 2026-04-29 & \nolinkurl{mazu-PrayerTracker-experiment-record-2026-04-29.md} \\
D2 & HandTracker experiment record, 2026-04-30 & \nolinkurl{mazu-HandTracker-experiment-record-2026-04-30.md} \\
D3 & Gesture-recognition optimization history and parameter-tuning notes & \nolinkurl{gesture-recognition-optimization-history-and-parameter-tuning-notes.md} \\
D4 & Gesture-recognition repair implementation plan & \nolinkurl{gesture-recognition-repair-implementation-plan.md} \\
D5 & Post-repair performance assessment for gesture recognition & \nolinkurl{post-repair-performance-assessment-for-gesture-recognition.md} \\
D6 & Prompt synchronization diagnosis and repair plan, 2026-05-03 & \nolinkurl{prompt-synchronization-diagnosis-and-repair-plan-2026-05-03.md} \\
D7 & Interface-level gesture optimization plan, version 01 & \nolinkurl{interface-level-gesture-optimization-plan-v01.md} \\
D8 & Mazu gesture-interaction revision notes & \nolinkurl{mazu-gesture-interaction-revision-notes.md} \\
\bottomrule
\end{tabularx}
\end{table}

\end{document}